\documentclass[accepted]{uai2023} 

\usepackage[british]{babel}

\usepackage{natbib} 
\usepackage{mathtools} 
\usepackage{booktabs}  
\usepackage{tikz} 

\usepackage{amsmath}
\usepackage{amssymb}
\usepackage{amsfonts}
\usepackage{amstext}
\usepackage{amsthm}
\usepackage{algorithm}
\usepackage{algorithmicx}
\usepackage{algpseudocode}
\usepackage{xcolor}
\usepackage{bbm}
\usepackage{mathtools}

\newcommand{\bx}{\mathbf{x}}

\DeclareMathOperator*{\argmax}{arg\,max}

\newtheorem{theorem}{Theorem}

\newtheorem{lemma}{Lemma}

\title{Benefits of Monotonicity in Safe Exploration with Gaussian Processes}

\author[1]{\href{mailto:<arpan@u.nus.edu>?Subject=Your UAI 2023 paper}{Arpan~Losalka}{}}
\author[1,2,3]{Jonathan~Scarlett}
\affil[1]{%
    Department of Computer Science\\
    National University of Singapore\\
    Singapore
}
\affil[2]{%
    Department of Mathematics\\
    National University of Singapore\\
    Singapore
}
\affil[3]{%
    Institute of Data Science\\
    National University of Singapore\\
    Singapore
}
  
\begin{document}
\maketitle

\begin{abstract}
  We consider the problem of sequentially maximising an unknown function over a set of actions while ensuring that every sampled point has a function value below a given safety threshold. We model the function using kernel-based and Gaussian process methods, while differing from previous works in our assumption that the function is monotonically increasing with respect to a \emph{safety variable}. This assumption is motivated by various practical applications such as adaptive clinical trial design and robotics.  Taking inspiration from the \textsc{\sffamily GP-UCB} and \textsc{\sffamily SafeOpt} algorithms, we propose an algorithm, monotone safe {\sffamily UCB} (\textsc{\sffamily M-SafeUCB}) for this task.  We show that \textsc{\sffamily M-SafeUCB} enjoys theoretical guarantees in terms of safety, a suitably-defined regret notion, and approximately finding the entire safe boundary. In addition, we illustrate that the monotonicity assumption yields significant benefits in terms of the guarantees obtained, as well as algorithmic simplicity and efficiency. We support our theoretical findings by performing empirical evaluations on a variety of functions, including a simulated clinical trial experiment.
\end{abstract}

\section{Introduction}\label{sec:intro}
The sequential optimisation of an unknown and expensive-to-evaluate function $f$ is a fundamental task with a number of interesting challenges. This task arises in various real-world applications, such as robotics \citep{lizotte2007automatic}, hyperparameter tuning in machine learning \citep{snoek2012practical}, environmental monitoring \citep{srinivas2012information}, adaptive clinical trial design \citep{takahashi2021mtd}, recommendation systems \citep{vanchinathan2014explore}, and many others. Gaussian process (GP) based techniques such as \textsc{\sffamily GP-UCB} \citep{srinivas2012information}, Thompson Sampling \citep{thompson1933likelihood} and Expected Improvement \citep{mockus1978application} are particularly popular for this task. 

In recent years, a variety of works have considered the important issue of \emph{safety}, where some actions (function inputs) need to be avoided altogether. Various algorithms such as \textsc{\sffamily SafeOpt} \citep{sui2015safe}, \textsc{\sffamily StageOpt} \citep{sui2018stagewise} and \textsc{\sffamily SafeOpt-MC} \citep{berkenkamp2021bayesian} have been proposed to tackle this problem. The main idea behind these algorithms is to start with a safe seed set of inputs, and sequentially expand and explore the candidate set of potentially safe points to eventually identify a \emph{reachable safe set} and/or the maximiser within that set. 

In this work, our main goal is to show that \emph{monotonicity with respect to just a single input variable} can be highly beneficial for this task.  Consider a function $f$ that we would like to maximise while ensuring that all selected points have value at most $h$.\footnote{This is distinct from previous works that require a value of \emph{at least} $h$, and we discuss the differences in Section \ref{sec:problem}.}
We assume that the unknown function $f$ is monotonically increasing (not necessarily strictly increasing) with respect to a \emph{safety variable} $s \in [0,1]$, while possibly remaining highly non-monotone with respect to the remaining variables $\mathbf{x}$. We consider performance measures based on both a form of cumulative regret and a notion of identifying the entire safe region. Briefly, the benefits of monotonicity in $s$ are:
\begin{itemize} \itemsep0ex
    \item[(i)] Our theoretical bounds have improved dependencies over previous works, particularly with respect to the domain size (see Appendix C.1 for details).
    \item[(ii)] By exploiting the monotonicity, we can circumvent the need to explicitly keep track of potential expanders, as existing algorithms do.
    \item[(iii)] Under monotonicity, continuity, and the mild  additional assumption that $s=0$ is always safe, we show that every safe point is reachable, which is not the case for general non-monotone functions.
\end{itemize}
Intuitively, the presence of the safety variable allows the algorithm to choose how cautious it should be while exploring the domain, and creates a more favourable function landscape for exploration.  For instance, the algorithm can ``back off'' or ``proceed with caution'' (lower $s$) when considering less-explored $\mathbf{x}$ values, but subsequently act more aggressively (higher $s$) when it becomes more confident that it is safe to do so.

\paragraph{Applications:} Consider the task of adaptive clinical trial design, where the goal is to recruit patients for drug trials in order to evaluate the safety and efficacy of a drug or drug combinations \citep{coffey2008adaptive, berry2006bayesian}.  
It is well-accepted that patient characteristics play a significant role in both safety and efficacy given a drug dose \citep{lee2020contextual}. Thus, it is helpful to model the drug dose as the safety variable $s$, and patient characteristics as the variable $\mathbf{x}$, both of which influence the unknown toxicity, say $f_{tox}$. For many classes of drugs such as cytotoxic agents, the toxicity and efficacy both increase strictly as the drug dose is increased \citep{chevret2006statistical}. In such cases, for Phase I clinical trials, it is usually necessary to find the \textit{Maximum Tolerated Dose (MTD)}, which is the dose with the maximum toxicity within the permitted threshold for the patient characteristics under consideration \citep{aziz2021multi, riviere2014bayesian, shen2020learning}.  Formulating this task in our problem setting can result in (i) maximisation of benefits (via regret minimisation) and (ii) minimisation of harmful effects to patients involved in the study (via safety constraints), while simultaneously (iii) identifying safety information about the entire set of patient characteristics (via sub-level set estimation), each of which are important goals of adaptive clinical trial design.  GP optimisation has been recently used for this task \citep{takahashi2021mtd, takahashi2021bayesian}, but the safety constraints were met in these works by being highly cautious in dose increments (single step increments for discrete dosage levels), and no theoretical guarantees were sought.

Problems in robotics may also serve as potential applications for our problem setup. For example, consider the scenario where a robot performs a task with certain parameters given by the variable $\mathbf{x}$, but that there also exists a parameter $s$ indicating the \emph{speed} (or more generally, any measure of ``caution'' with lower values being more cautious) at which the task is attempted.  Then, one may seek to optimise the parameters while ensuring that $s$ is never pushed too high to become unsafe, leading to a natural monotonicity constraint. We explore a simple inverted pendulum problem of this kind in Section \ref{sec:exps}. We note that in some cases, it may be more natural to have separate functions $f$ and $g$ for measuring reward and safety, and we discuss such variations in Section \ref{sec:algo}.

\paragraph{Related Work:}
\citet{sui2015safe} proposed the first algorithm, \textsc{\sffamily SafeOpt}, for safe GP optimisation. \textsc{\sffamily StageOpt} was proposed by \citet{sui2015safe} as a variation of \textsc{\sffamily SafeOpt}, where safe set finding and function optimisation were separated into two distinct phases. \citet{berkenkamp2021bayesian} proposed a generalised version of \textsc{\sffamily SafeOpt} called \textsc{\sffamily SafeOpt-MC} to tackle the problem when safety functions are decoupled from the function being optimised. Other algorithms such as GOOSE \citep{turchetta2019safe} seek to be more goal-directed during safe set expansion. Safe exploration using Gaussian processes has also been considered by \citet{schreiter2015safe}, but for the goal of active learning of the unknown function. To our knowledge, none of these works have explored the benefits of having a safety variable leading to monotonicity.  Moreover, their theoretical guarantees exhibit certain weaknesses with respect to the domain size that we are able to circumvent; see Section \ref{sec:theory} for the details.

Another related line of work introduces algorithms such as \textsc{\sffamily GoSafe} \citep{Baumann2021GoSafeGO} and \textsc{\sffamily GoSafeOpt} \citep{Sukhija2022ScalableSE} to address safe Bayesian optimisation problems.  Their setting is fundamentally different to ours, since they consider a \emph{dynamic} system with time-varying inputs, and allow \emph{intervening} with a safe backup policy when an imminent safety violation is detected in the original policy.  Our setup considers static functions without interventions (which may not be feasible in some applications, e.g., once a drug dose is administered it may not be possible to change).

A different approach to safe GP optimisation is taken in \citep{amani2021regret}, in which conditions are explored under which an initial safe seed set can be sampled enough times for the resulting samples alone to expand the safe set significantly (and include the global safe maximiser).  However, this requires careful assumptions on the seed set depending strongly on the kernel, and the idea appears to be most suited to finite-dimensional feature spaces (e.g., linear or polynomial kernels); see Appendix C.3 for discussion.

In a parallel line of work, the problem of level set estimation and related settings involving excursion sets has been considered, e.g., see \citep{gotovos2013active,bogunovic2016truncated, bolin2015excursion}.  GP optimisation with monotonicity assumptions has also been considered by \citet{li2017bayesian} and \citet{wang2018bayesian}.  However, these works do not consider safety constraints, and consequently, the associated algorithms are significantly different.

A notable prior work combining safety and monotonicity is \citep{wang2022best}, but they study a non-GP setting where all the arms (corresponding to $\bx$ in our setting) are modeled separately, and the goal is best-arm identification.  This leads to a precise characterisation of the number of arm pulls. However, their setup, algorithm, and results remain very different from our work, where smoothness with respect to $\bx$ (as well as $s$) plays a crucial role.

Finally, safety has been considered in a variety of other settings including linear bandits  \citep{amani2019linear,khezeli2020safe} and reinforcement learning \citep{turchetta2016safe,berkenkamp2017safe,turchetta2020safe}, but compared to the works outlined above for GP settings, these are less directly relevant to ours.

\paragraph{Contributions:}
Summarising the above discussions, we list our main contributions as follows:
\begin{itemize}
    \item We study the problem of safe sequential optimisation of an unknown function, and introduce the idea of considering monotonicity of the function with respect to a ``safety'' variable. We propose the monotone safe {\sffamily UCB} (\textsc{\sffamily M-SafeUCB}) algorithm for this problem.
    \item We show that with high probability, \textsc{\sffamily M-SafeUCB} achieves sub-linear regret (for a suitably-defined regret notion to follow), only selects safe actions, and identifies the safe (sub-level) set of the function with high accuracy.
    \item We experimentally evaluate \textsc{\sffamily M-SafeUCB} alongside other baselines on a variety of functions, and demonstrate that the resulting performance aligns with the theoretical guarantees.
\end{itemize}

\section{Problem Statement} \label{sec:problem}

We consider the problem of sequentially maximising a fixed but unknown function $f: \mathcal{D} \rightarrow \mathbb{R}$ over a set of decisions while satisfying safety constraints, where $\mathcal{D} = \mathcal{D_S} \times \mathcal{D_X}$, $\mathcal{D_X} \subset \mathbb{R}^d$ is a compact set and $\mathcal{D_S} = [0,1]$.  As discussed above, we assume that the function is monotonically increasing in the first argument. At each round $t$, an algorithm selects an \textit{action} $(s_t, {\mathbf{x}}_t) \in \mathcal{D_S} \times \mathcal{D_X}$, and subsequently observes the \textit{noisy reward} $y_t = f(s_t, {\mathbf{x}}_t) + \epsilon_t $. The action must be chosen at round $t$ such that it depends upon the actions picked and the rewards observed up to round $t-1$, denoted by $\mathcal{H}_{t-1} = \{(s_k, {\mathbf{x}}_k, y_k) : k = 1, \dotsc, t-1\}$ (i.e., the history). The algorithm is also required to satisfy the safety constraint $f(s_t, \mathbf{x}_t) \leq h \; \forall t \geq 1$ with high probability. 

\paragraph{Goal:} The goals of an algorithm in our problem setting include maximising its cumulative reward and/or finding the entire $h$-sub-level set of $f$, while only choosing safe actions. These desiderata are formalised as follows.  For cumulative regret, we consider the following definition:
\begin{equation}
    R_T = \sum_{t=1}^T r_t \text{, where } r_t = h - f(s_t, \mathbf{x}_t),
\end{equation}
where we compare against $h$ rather than $\max_{s, \bx \in \mathcal{D}} f(s, \bx)$ in view of the safety requirement.  For the sub-level set, we define 
\begin{equation}
    L_h(f) = \{(s, \mathbf{x}) \in \mathcal{D} | f(s, \mathbf{x}) \leq h\},
\end{equation}
which we seek to approximate to high accuracy (see below).  For the safety requirement, we seek that the sampled points satisfy $f(s_t, \mathbf{x}_t) \leq h \; \forall t \geq 1$ with high probability.

Returning to the notion of the safe sub-level set, we quantify the quality of a solution $\hat{L}$ returned by an algorithm after $T$ rounds, with respect to a given point $(s,\mathbf{x}) \in \mathcal{D}$, using the following misclassification loss:
\begin{equation}
\label{eqn:misclassification}
l_h(s,\mathbf{x}) = 
    \begin{cases}
      \max\{0,&\negthickspace h- f(s,\mathbf{x})\} \text{ if } (s,\mathbf{x}) \notin \hat{L}, \\
          \infty, & \text{if } (s,\mathbf{x}) \in \hat{L} \text{ and } (s,\mathbf{x}) \notin L_h(f), \\ 
      0, & \text{if } (s,\mathbf{x}) \in \hat{L} \text{ and } (s,\mathbf{x}) \in L_h(f).     \end{cases}
\end{equation}
Essentially, this loss function penalises an algorithm heavily for classifying unsafe points as safe, while the penalty for classifying safe points as unsafe increases linearly with the difference in the function value from the threshold. We require that the algorithm should return an \textit{$\epsilon$-accurate solution} with probability at least $1-\delta$, i.e., 
\begin{equation}
\mathbb{P}\left\{\max_{s,\mathbf{x} \in \mathcal{D}} l_h(s,\mathbf{x}) \leq \epsilon \right\} \geq 1-\delta.  \label{eq:level_set_whp}
\end{equation}

We note that the notion of regret that we consider is primarily of interest \emph{when coupled with \eqref{eq:level_set_whp}}, rather than in itself.  Small $R_T$ is generally desirable since it implies that we are eventually sampling points with the highest possible safe function value.  However, one way of achieving small $R_T$ might be to always choose the same $\mathbf{x}$ and gradually increasing $s$ until a low-regret point is found.  The additional condition \eqref{eq:level_set_whp} precludes this possibility.

\paragraph{Assumptions:} Certain smoothness assumptions on the function $f$ are necessary in order to be able to provide theoretical guarantees. Similar to much of the earlier work in the area of GP optimisation, we assume that $f$ has bounded norm in the reproducing kernel Hilbert space (RKHS) of functions $\mathcal{D} \rightarrow \mathbb{R}$, with positive semi-definite kernel function $k: \mathcal{D} \times \mathcal{D} \rightarrow \mathbb{R}$. This RKHS, denoted by $\mathcal{H}_k(\mathcal{D})$, is completely specified by its kernel function $k(\cdot, \cdot)$ and vice-versa, with an inner product $\langle\cdot, \cdot\rangle_k$ obeying the reproducing property: $f(\mathbf{z}) = \langle f, k(\mathbf{z}, \cdot)\rangle_k \; \forall f \in \mathcal{H}_k(\mathcal{D})$. The RKHS norm ${\vert\vert f \vert\vert}_k = \sqrt{{\langle f,f \rangle}_k}$ is a measure of the smoothness of $f$ with respect to the kernel function $k$, and satisfies $f \in \mathcal{H}_k(\mathcal{D})$ if and only if ${\vert\vert f \vert\vert}_k < \infty$. We assume a known upper bound $B$ on the RKHS norm of the unknown target function, i.e., $||f||_k \leq B$. We also adopt the standard assumption of bounded variance: $k(\mathbf{z}, \mathbf{z}) \leq 1 \; \forall \mathbf{z} \in \mathcal{D}$. 

In addition, we make the following assumptions regarding the function domain, monotonicity, and safety:
\begin{enumerate}
    \item $\mathcal{D_S} = [0,1]$ is continuous, while $\mathcal{D_X}$ can be either discrete or continuous (our algorithms are written for the discrete case, and we discuss the distinction between the two in Appendix C.2);
    \item The function $f$ is monotonically increasing in the first argument, i.e., for all $\mathbf{x} \in \mathcal{D_X}$, $f(s,\mathbf{x})$ is a non-decreasing function of $s \in \mathcal{D_S}$;
    \item The action $(0, \mathbf{x})$ is safe for every $\mathbf{x}$ in the domain, i.e., for all $\mathbf{x} \in \mathcal{D_X}$, $f(0,\mathbf{x}) \leq h$;
    \item The function $f$ exceeds the threshold $h$ for at least one point in the domain, i.e., $\max_{(s, \mathbf{x} \in \mathcal{D})} f(s, \mathbf{x}) > h$.
\end{enumerate}
The third assumption above is natural since $s=0$ corresponds to the most cautious selection possible, and the fourth assumption is mild since otherwise every action is safe and hence no algorithm would ever choose an unsafe action.

Finally, the noise sequence $\{\epsilon_t\}_{t \geq 1}$ is assumed to be conditionally $R$-sub-Gaussian for a fixed constant $R \geq 0$, i.e., 
\begin{equation}
\label{eqn:noise}
    \forall t \geq 0, \forall \lambda \in \mathbb{R}, \mathbb{E}\left[e^{\lambda \epsilon_t} \vert \mathcal{F}_{t-1}\right] \leq \exp\left(\frac{\lambda^2 R^2}{2}\right)
\end{equation}
where $\mathcal{F}_{t-1}$ is the $\sigma$-algebra generated by the random variables $\{s_k, \mathbf{x}_k, \epsilon_k\}_{k=1}^{t-1}$ and $\mathbf{x}_t$. 

\paragraph{Difference to Existing Settings:} An important distinction between our work and certain previous ones (e.g., \citep{sui2015safe,sui2018stagewise}) is that we consider all points \emph{below} the threshold to be safe, rather than all points above the threshold.  Our setting corresponds to trying to maximise a function while avoiding the risk of pushing it too far (e.g., dosage in clinical trials), whereas the alternative setting corresponds to needing to avoid excessively low-performance decisions (e.g., parameter configurations that may cause a drone to crash).  The resulting algorithms are somewhat different since in our setting, maximisation algorithms will have a natural tendency to move closer to the safety threshold.

While the above difference is important to keep in mind, it is also worth noting that it becomes insignificant when one considers variations of the problem with \emph{separate} safety and reward functions (see Section \ref{sec:algo} for further discussion).  In addition, in our experiments we adapt \textsc{\sffamily SafeOpt} to suit our setting.

\section{Proposed Algorithm} \label{sec:algo}
\paragraph{Gaussian Process Model:} As is common in prior works, we consider algorithms that use Bayesian modeling (despite the non-Bayesian problem formulation).  For this purpose, we use a Gaussian likelihood model for the observations, and a Gaussian process (GP) prior for uncertainty over the unknown function $f$.  We let $GP(\mu(\cdot), k(\cdot, \cdot))$ denote a GP with mean $\mu$ and kernel $k$.  In the following, we often shorten the GP input $(s, \mathbf{x})$ to $\mathbf{z}$ to simplify notation.

The algorithm uses a zero-mean GP, $GP(0, k(\cdot, \cdot))$, with $k$ being the same as that defining the RKHS.  The Gaussian likelihood has an associated variance parameter, which we denote by $\lambda$ (i.e., corresponding to additive $\mathcal{N}(0,\lambda)$ noise in the Bayesian model).

With the Bayesian model in place, we have the following standard posterior update equations:
\begin{align}
\mu_t(\mathbf{z}) &=k_t(\mathbf{z})^T\left(K_t+\lambda I\right)^{-1} \mathbf{y}_{t}, \\
k_t\left(\mathbf{z}, \mathbf{z}^{\prime}\right) &=k\left(\mathbf{z}, \mathbf{z}^{\prime}\right)-k_t(\mathbf{z})^T\left(K_t+\lambda I\right)^{-1} k_t\left(\mathbf{z}^{\prime}\right), \\
\sigma_t^2(\mathbf{z}) &=k_t(\mathbf{z}, \mathbf{z}).
\end{align}

\paragraph{Proposed Algorithm:} We propose an algorithm called monotone safe {\sffamily UCB} (\textsc{\sffamily M-SafeUCB}), and provide its theoretical guarantees in Section \ref{sec:theory}.  
The key idea is to exploit the knowledge that the function $f$ is monotonically increasing in the first argument, $s$, and thus, continually sample points in the domain that have their upper confidence bound equal to the threshold value $h$. 

In more detail, \textsc{\sffamily M-SafeUCB} uses a (standard) combination of the current posterior mean and standard deviation to construct an upper confidence bound (UCB) envelope for the function $f$ over $\mathcal{D}$, given by
\begin{equation}
    \mathrm{UCB}_{t-1}(s, \mathbf{x}) = \mu_{t-1}(s, \mathbf{x})  + \beta_t \sigma_{t-1}(s, \mathbf{x}), \label{eq:ucb}
\end{equation}
where $\beta_t$ is a time-dependent constant, that is set as per Lemma \ref{lemma:beta} below. In each round $t$, it chooses a sample such that $\mathrm{UCB}_{t-1}(s, \mathbf{x})= h$ (favouring higher $s$ in the rare case of having multiple such $s$ for a single $\mathbf{x}$). This trades off between exploration and exploitation, i.e., it leads to selection of more points close to the currently optimal solution while exploring as well. If multiple such points are available with UCB equal to $h$, then \textsc{\sffamily M-SafeUCB} selects the one that has the maximum posterior variance, thus helping to reduce uncertainty and encourage exploration.

\begin{algorithm}[!t]
\caption{\textsc{\sffamily M-SafeUCB}}\label{alg:monotone_UCB}
\begin{algorithmic}
\State \textbf{Input:} Prior $GP(0,k)$, parameters $R, B, \lambda, \delta$ 
\For{$t = 1, \dotsc, T$}
    \State $\beta_t = B+R \sqrt{2\left(\gamma_{t-1}+1+\ln (1 / \delta)\right)} \quad $ 
    \State $S_t = \emptyset$
    \For{$\mathbf{x} \in \mathcal{D_X}$}    \Comment{{\small find max. safe $s \, \forall \mathbf{x}$}}
        \If{$ \mathrm{UCB}_{t-1}(s, \mathbf{x}) > h \,\forall s:(s, \mathbf{x}) \in \mathcal{D}$}
            \State $s_t^{(\mathbf{x})} = 0$
            \State $S_t = S_t \cup \{ (s_t^{(\mathbf{x})}, \mathbf{x})\}$
        \ElsIf{$ \exists s:(s, \mathbf{x}) \in \mathcal{D} \text{, } \mathrm{UCB}_{t-1}(s, \mathbf{x}) = h $}
            \State{$
            \begin{aligned}
                s_t^{(\mathbf{x})} = \max\{s : (s, & \mathbf{x}) \in \mathcal{D}, \\ & \mathrm{UCB}_{t-1}(s, \mathbf{x}) = h \}
            \end{aligned}
            $}
            \State $S_t = S_t \cup \{ (s_t^{(\mathbf{x})}, \mathbf{x})\}$
        \EndIf
    \EndFor
    \If{$S_t = \emptyset$}  \Comment{{\small if safe everywhere, set $s=1 \, \forall \bx$}}
        \For{$\mathbf{x} \in \mathcal{D_X}$}
            \State $S_t = S_t \cup \{ (1, \mathbf{x})\}$
        \EndFor
    \EndIf
    \State $(s_t, \mathbf{x}_t) = \argmax_{s, \mathbf{x} \in S_t} \sigma_{t-1}(s, \mathbf{x})$
    \State Update posterior to get $\mu_t, \sigma_t$
\EndFor
\State $
\begin{aligned}
\overline{\mathrm{UCB}}_T(s, \mathbf{x}) =  \min_{1 \leq t \leq T}  \mathrm{UCB}_{t} & (s, \mathbf{x}) \, \forall (s, \mathbf{x}) \in \mathcal{D}
\end{aligned}
$ 

\For{$\mathbf{x} \in \mathcal{D_X}$}    \Comment{{\small form safe boundary}}
    \If{$ \overline{\mathrm{UCB}}_T(s, \mathbf{x}) > h \,\forall s:(s, \mathbf{x}) \in \mathcal{D} $}
    \State $\overline{s}^{(\mathbf{x})}_{T} = 0$
    \Else    
    \State{
    $\begin{aligned}
			\overline{s}^{(\mathbf{x})}_{T} = \max\{s : & (s, \mathbf{x}) \in \mathcal{D},
			& \overline{\mathrm{UCB}}_T(s, \mathbf{x}) \leq h\}
	\end{aligned}$}
    \EndIf
\EndFor 
\State $\hat{L}_{T} = \{(s, \mathbf{x}) \in \mathcal{D}: s \leq \overline{s}^{(\mathbf{x})}_{T}\}$ \\
\Return $\hat{L}_{T}$
\end{algorithmic}
\end{algorithm}

To account for all possibilities that may arise in each round $t$, we set the candidate $s^{(\mathbf{x})}_t \in \mathcal{D_S}$ and the candidate set $S_t^{(\mathbf{x})}$ for each $\mathbf{x} \in \mathcal{D_X}$ as follows:
\begin{itemize}
    \item If there exists $(s,\mathbf{x}) \in \mathcal{D}$ such that $\mathrm{UCB}_{t-1}(s, \mathbf{x}) = h$, then $s_t^{(\mathbf{x})} = \max\{s : (s, \mathbf{x}) \in \mathcal{D}, \mathrm{UCB}_{t-1}(s, \mathbf{x}) = h \}$ and $S_t^{(\mathbf{x})} = \{(s_t^{(\mathbf{x})}, \mathbf{x})\}$;
    \item If $\forall s \in \mathcal{D_S}$, $\mathrm{UCB}_{t-1}(s, \mathbf{x}) > h$, then $s_t^{(\mathbf{x})} = 0$ (based on the assumption that for all $\mathbf{x} \in \mathcal{D_X}$,  $f(0,\mathbf{x}) \leq h$) and $S_t^{(\mathbf{x})} = \{(s_t^{(\mathbf{x})}, \mathbf{x})\}$;
    \item If $\forall s \in \mathcal{D_S}$, $\mathrm{UCB}_{t-1}(s, \mathbf{x}) < h$, then $S_t^{(\mathbf{x})} = \emptyset$.
\end{itemize}
Next, the set $S_t$ is formed by taking the union over all candidate sets $S_t^{(\mathbf{x})}$, i.e., $S_t = \bigcup_{\mathbf{x} \in \mathcal{D_X}} S_t^{(\mathbf{x})}$. Then, $(s_t, \mathbf{x}_t)$ is chosen by maximising the predictive variance:
\begin{equation}
    (s_t, \mathbf{x_t}) = \argmax_{(s, \mathbf{x}) \in S_t} \sigma_{t-1}(s, \mathbf{x}).
\end{equation}
We note that if no $(s, \mathbf{x}) \in \mathcal{D}$ satisfies $\mu_{t-1}(s, \mathbf{x})  + \beta_t \sigma_{t-1}(s, \mathbf{x}) \geq h$ for a certain $t \leq T$, then it must be the case that with high probability, the entire function $f$ lies below the safety threshold.  This is precluded by assumption 4 of our problem statement. However, with a low probability, it may also be the case that the noisy observations make the function ``appear'' to be below the threshold $h$ to the algorithm. In this low probability scenario, $\forall \mathbf{x} \in \mathcal{D_X}$, we set $s_t^{(\mathbf{x})} = 1$ and select $(s_t, \mathbf{x}_t) \in S_t$ as earlier by maximising variance.

While we make the mild assumption that at least one unsafe point exists, if it happens that $f$ is safe for all $s,\bx \in \mathcal{D}$, the algorithm will still eventually identify $s=1$ as safe for every $\bx$. In this scenario, the problem essentially becomes an unconstrained optimisation problem, and our notion of regret is no longer appropriate (since every point has a strict gap to $h$).

Finally, at the end of $T$ rounds, the algorithm considers the intersection of the confidence regions across all rounds and all $\mathbf{x} \in \mathcal{D_X}$, and forms an estimate of the safe sub-level set with respect to the upper bound of this intersection (denoted by $\overline{\mathrm{UCB}}_T(s, \mathbf{x})\}$) as follows:
\begin{align}
\overline{s}^{(\mathbf{x})}_{T} &= \max\{s : (s, \mathbf{x}) \in \mathcal{D},  \notag  \\ 
& \qquad \qquad \overline{\mathrm{UCB}}_T(s, \mathbf{x}) \leq h\}, \, \forall \mathbf{x} \in \mathcal{D_X}, \label{eqn:s_x_t}\\
\hat{L}_{T} &= \{(s, \mathbf{x}) \in \mathcal{D}: s \leq \overline{s}^{(\mathbf{x})}_{T}\}.
\end{align}

\paragraph{Note on Two-Function Settings:} Throughout the paper, we focus on the case that the function $f$ dictates both the objective (higher is better) and the safety (too high is unsafe).  However, our ideas can also be applied in scenarios where these functions differ; say, with $f(s,\mathbf{x})$ being the objective and $g(s,\mathbf{x})$ dictating the safety.  In the following, suppose that whenever we query a point, we observe noisy samples from both $f$ and $g$.  The two functions both have RKHS norm at most $B$, and the algorithm can form two separate GP posteriors for them.

First suppose that both $f$ and $g$ are monotone with respect to $s$.  Moreover, similar to the current setup, suppose that the objective is to find the highest possible $s$ associated with each $\mathbf{x}$.  Then, one can simply apply our main algorithm to $g$, with Theorem \ref{theorem:safe_boundary} guaranteeing that we (approximately) find the entire safe boundary.  Since both $f$ and $g$ are monotone, the highest safe $s$ for $g$ is also the highest safe $s$ for $f$, and the overall problem is essentially unchanged compared to the single-function setting. As an example, this scenario corresponds to the task of finding the \textit{MTD} in clinical trials as discussed in section \ref{sec:intro}.

In general, one may be interested in scenarios where only $g$ is monotone with respect to $s$, whereas $f$ is more general.  Even in such cases, Algorithm \ref{alg:monotone_UCB} could be used as an initial step to find the safe boundary of $g$, as is guaranteed by Algorithm \ref{alg:monotone_UCB}.  Then, the safe boundary could be passed to a downstream optimiser that seeks to maximise $f$ (either over all safe $(s,\bx)$, or over all $s$ for each individual $\bx$).  This is akin to how the stage-wise algorithm in \citep{sui2018stagewise} operates.

Finally, more sophisticated algorithms may be possible that utilise information observed about $f$ and $g$ jointly throughout the course of the algorithm, but such investigations are left for future work.

\paragraph{Note on Contextual Settings:} In some scenarios, it may be useful to incorporate \textit{context variable} $\mathbf{c}$ in the problem setup, besides the \textit{action variables} $s,\bx$.  For example, a clinical trial setting different from our previous motivating one might consist of choosing a drug dose $s$ (with monotone behavior) and the dosage of a different drug $\bx$ (not necessarily monotone), while also having access to patient characteristics $\mathbf{c}$ (e.g., BMI).  All of these impact the drug toxicity, but in contrast to $(s,\bx)$, it may be that $\mathbf{c}$ cannot be chosen actively and is instead given by ``nature''.

In such scenarios, Algorithm \ref{alg:monotone_UCB} can be readily extended to incorporate context variables following the discussion on contextual safe Bayesian optimisation from \citep{berkenkamp2021bayesian}. This approach allows joint modelling of the unknown function with the variables $(s,\bx, \mathbf{c})$, so that information can be shared across contexts during optimisation. The proposed selection rule in this case would be: \textit{for the given context} $\mathbf{c}_t$, search through the $\bx$'s and find the highest possible safe $s$ associated with each $\bx$ (using the $\mathrm{UCB}$ as usual), and then among all candidate $(s,\bx)$'s, choose $(s_t,\bx_t)$ to be the one with the highest posterior variance. Since the idea behind this extension closely follows \citep{berkenkamp2021bayesian}, we do not elaborate on it further in this work.

\section{Theoretical Results} \label{sec:theory}

We present our main theoretical results in this section, under the set of assumptions outlined in \ref{sec:problem}. The proofs are provided in Appendix A. We note that in asymptotic statements, we treat the dimension $d$ and sub-Gaussian parameter $R$ (see \eqref{eqn:noise}) as constants, and similarly treat the kernel as fixed.

\begin{lemma}
\label{lemma:beta}
(Theorem 2, \citep{chowdhury2017kernelized}) Fix $\delta > 0$, and suppose that $\beta_t$ is set as follows: 
\begin{equation}
\beta_t = B+R\sqrt{2(\gamma_{t-1}+1+\ln{(1/ \delta)})}. \label{eq:beta_t}
\end{equation}
Then, we have the following with probability at least $1-\delta$:
\begin{equation}
    | \mu_{t-1} (s, \mathbf{x}) - f(s, \mathbf{x})| \leq  \beta_t \sigma_{t-1}(s, \mathbf{x}), \quad \forall \bx, s, t,
\end{equation}
where $\gamma_t$ is the maximum information gain at time $t$:
\begin{equation}
    \gamma_t := \max_{A \subset \mathcal{D}: |A| = t} I(y_A; f_A).
\label{eqn:gamma}
\end{equation}
Here, $I(y_A; f_A)$ denotes the mutual information between $f_A = [f(\mathbf{x})]_{x \in A}$ and $y_A = f_A + \epsilon_A$, where $\epsilon_A \sim \mathcal{N}(0, \lambda I)$. 
\end{lemma}

This lemma follows from Theorem 2 from \citep{chowdhury2017kernelized}.  The quantity $\gamma_t$ is ubiquitous in the GP bandit literature, and quantifies the maximum possible reduction in uncertainty about $f$ after observing $y_A$ at a set of points $A \subset \mathcal{D}$. 

\begin{theorem}
\label{theorem:regret_bound}
\textit{Under the setup and assumptions of Section \ref{sec:problem}, and the choice of $\beta_t$ in Lemma \ref{lemma:beta}, \textsc{\sffamily M-SafeUCB} satisfies the following regret bound with probability at least $1-\delta$:}
\begin{equation}
R_T = O\left(B\sqrt{T\gamma_T} + \sqrt{T\gamma_T(\gamma_T + \ln(1/\delta))}\right).
\end{equation}
\end{theorem}

As an example, with the squared exponential (SE) kernel on a compact subset $\mathcal{D} \subset \mathbb{R}^d$, $\gamma_T$ is $O({\ln^{d+1}{T}})$ (Srinivas et al., 2010). Thus $R_T/T \rightarrow 0 $ as $T \rightarrow \infty$, resulting in sub-linear regret.  The same holds for the Mat\'ern kernel when the smoothness parameter $\nu$ is not too small. 

The following theorem formalises the statement that the algorithm approximately identifies the entire safe region.

\begin{theorem}
\label{theorem:safe_boundary}
Consider the setup and assumptions of Section \ref{sec:problem}, and the choice of $\beta_t$ in Lemma \ref{lemma:beta}.  
With $l_h(s, \mathbf{x})$ as defined in equation \eqref{eqn:misclassification}, \textsc{\sffamily M-SafeUCB} finds an $\epsilon$-accurate solution $\hat{L}$ with probability at least $1-\delta$:
\begin{equation}
\mathbb{P}\left\{\max_{s,\mathbf{x} \in \mathcal{D}} l_h(s,\mathbf{x}) \leq \epsilon \right\} \geq 1-\delta,
\end{equation}
where $\epsilon$ scales as follows:
\begin{equation}
\epsilon = O\left(B\sqrt{\gamma_T/T} + \sqrt{(\gamma_T + \ln(1/\delta))\gamma_T/T}\right).
\end{equation}
\end{theorem}
Substituting the bound on $\gamma_T$ stated above for the squared exponential kernel (or the Mat\'ern kernel when $\nu$ is not too small), we find that $\epsilon \rightarrow 0$ as $T \rightarrow \infty$.

The proof of Theorem \ref{theorem:regret_bound} follows a similar general structure to regret analyses for \textsc{\sffamily GP-UCB} and related algorithms \citep{srinivas2012information,chowdhury2017kernelized}.  In contrast, Theorem \ref{theorem:safe_boundary} requires less standard techniques; see Appendix A for the complete argument.

\begin{figure*}[t!]
  \centering
  \setlength\tabcolsep{2pt}
  \begin{tabular}{ccc}
     \includegraphics[width=0.33\linewidth]{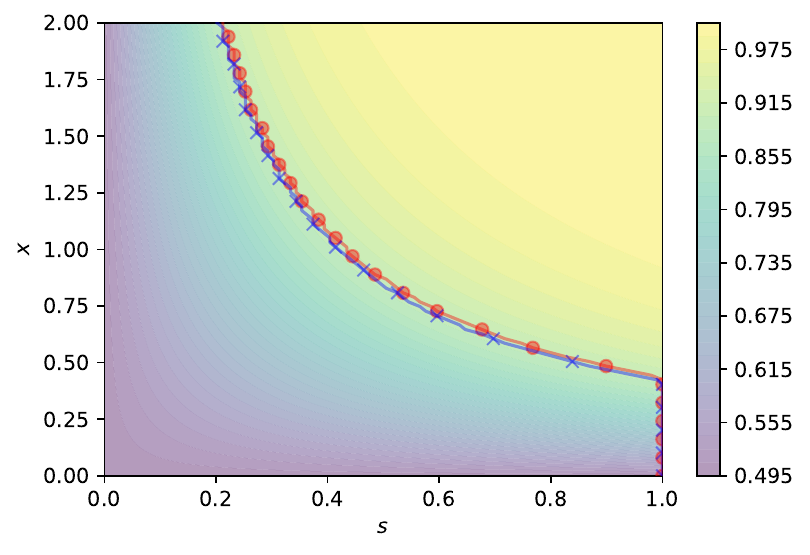} &
     \includegraphics[width=0.33\linewidth]{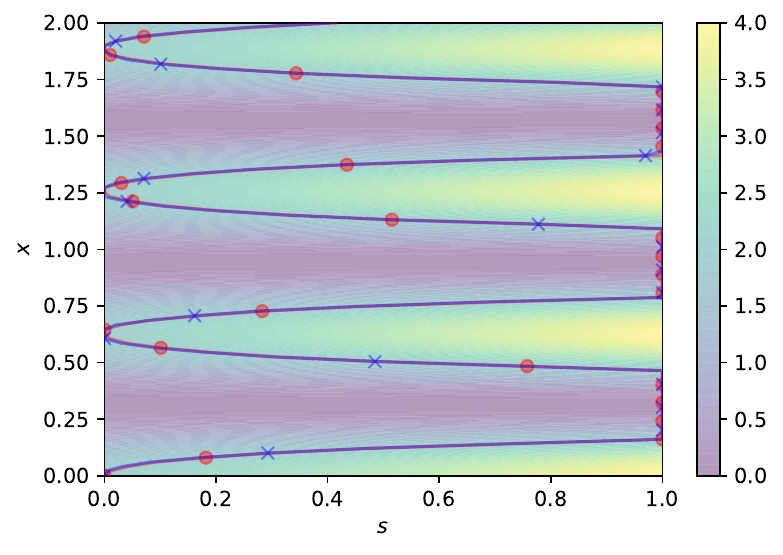} &
     \includegraphics[width=0.33\linewidth]{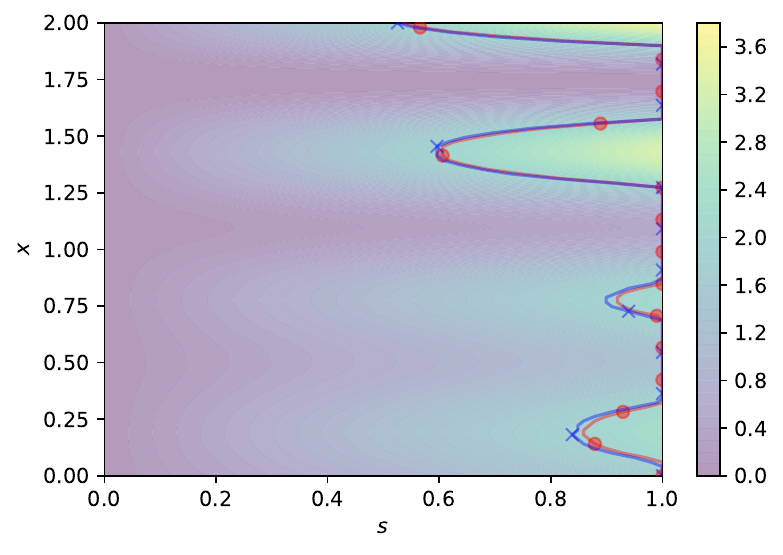} \\ 
     \includegraphics[width=0.33\linewidth]{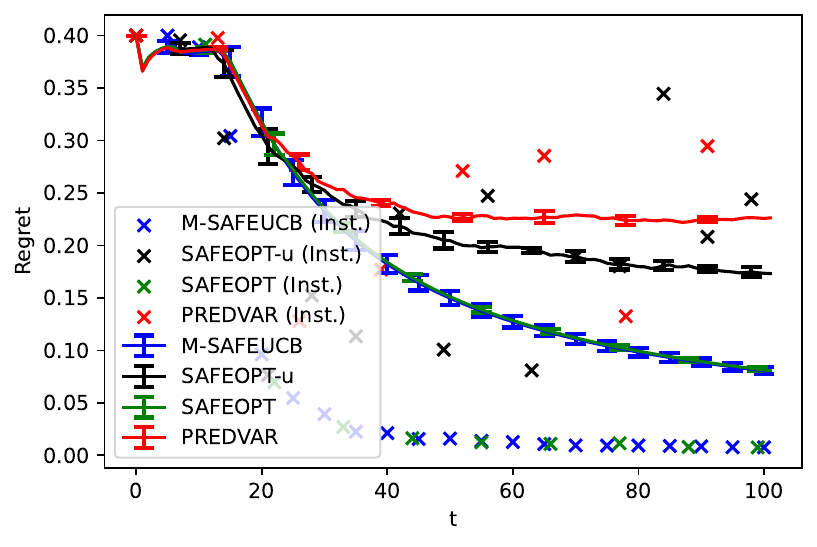} &
     \includegraphics[width=0.33\linewidth]{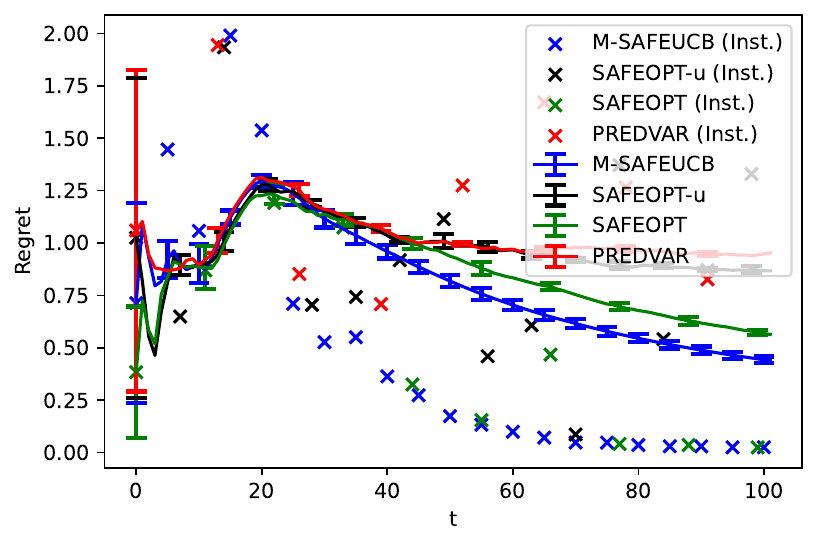} &
     \includegraphics[width=0.33\linewidth]{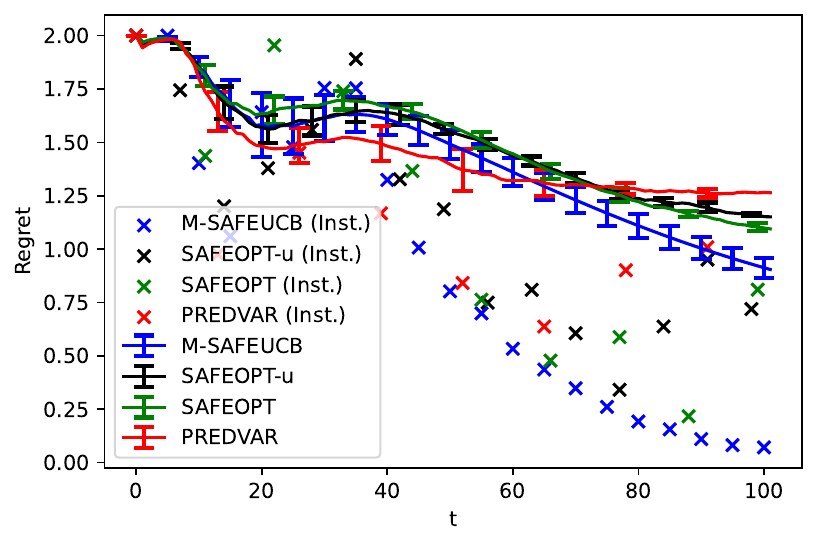} \\
      \includegraphics[width=0.33\linewidth]{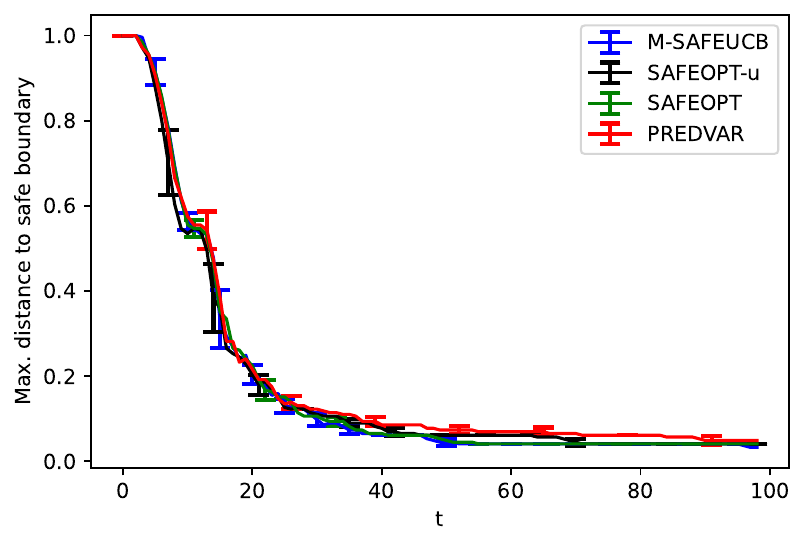} &
     \includegraphics[width=0.33\linewidth]{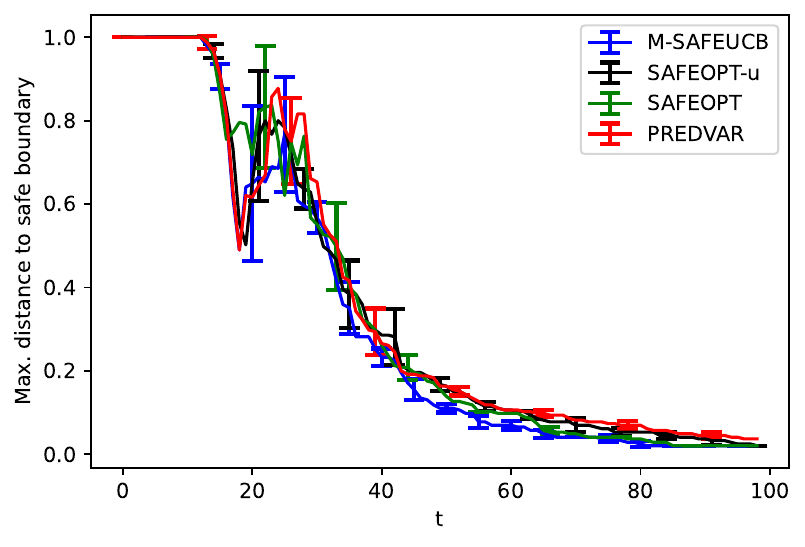} &
     \includegraphics[width=0.33\linewidth]{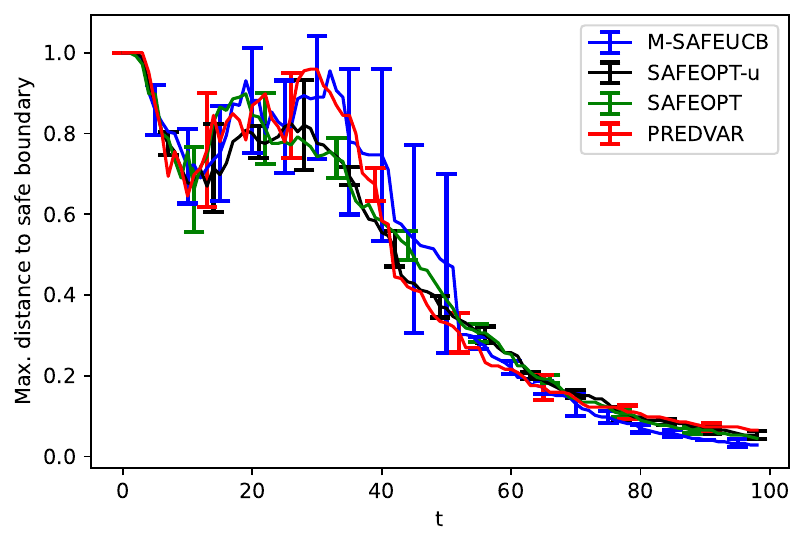} \\
     $f_{tox}(s, x)$ & $f_{syn_1}(s,x)$ & $f_{syn_2}(s,x)$
    \end{tabular}
  \caption{The results of running \textsc{\sffamily M-SafeUCB} on the functions $f_{tox}$ (first column), $f_{syn_1}$ (second column) and $f_{syn_2}$ (third column). All functions are monotonically increasing in the input variable $s$. The first row shows the \textit{safe boundary} as predicted by the algorithm (in blue) along with the actual safe boundary (in red), overlaid on the plot of the function. The second row shows the plots for the instantaneous and average cumulative regrets incurred by \textsc{\sffamily M-SafeUCB}, as well as \textsc{\sffamily SafeOpt} and \textsc{\sffamily PredVar} (\textsc{\sffamily SafeOpt}-u denotes using an underestimate of the Lipschitz constant). The last row shows the maximum distance to the safe boundary across $x \in [0,2]$ as a function of the time step. In each case, \textsc{\sffamily M-SafeUCB} is able to find the safe boundary almost exactly without sampling unsafe points, and the regret decreases towards zero.
  \label{fig:mgpucb_bo_safe_regret}}
\end{figure*}

\paragraph{Dependence on $\mathcal{D_X}$:} As we mentioned in Section \ref{sec:intro}, our theory circumvents strong dependencies on the domain size that are present in previous works for safe settings, and in fact holds even for continuous domains. The guarantees of \textsc{\sffamily SafeOpt} \citep{sui2015safe} (and related algorithms in follow-up works) roughly state that the entire reachable safe region, up to deviations of $\epsilon$, will be identified with high probability once the time horizon $T$ satisfies
\begin{equation}
    \frac{T}{\beta_T \gamma_T} \ge \frac{C |\bar{R}_0|}{\epsilon^2}, \label{eq:SafeOPT_T}
\end{equation} 
where $C$ is a suitably-defined constant, and $\bar{R}_0 \subseteq \mathcal{D_X}$ is a safe region that can potentially be reached from some initial safe seed set.  While their setup is slightly different from ours (see Section \ref{sec:problem}), the associated guarantees readily transfer without significant modification. Under the mild assumption that $\bar{R}_0$ occupies a constant fraction of the domain, the requirement in \eqref{eq:SafeOPT_T} incurs a linear dependence on the domain size. 

Our main results (Theorems \ref{theorem:regret_bound} and \ref{theorem:safe_boundary}) show that, in fact, the dependence on the domain size can be avoided altogether when we have the additional variable $s$ that the function is monotone with respect to. We refer the reader to Appendix C.1 for further discussion in this regard.

\paragraph{Dependence on $\gamma_T$ and $T$:} Our regret bound in Theorem \ref{theorem:regret_bound} incurs $\gamma_T \sqrt{T}$ dependence on $T$ (up to logarithmic terms), and our convergence rate in Theorem \ref{theorem:safe_boundary} analogously incurs dependence $\frac{\gamma_T}{\sqrt{T}}$. This dependence matches that of \textsc{\sffamily GP-UCB} \citep{srinivas2012information} and other related algorithms for the standard (non-safe) setting, as well as \textsc{\sffamily SafeOpt} (and others) for the safe setting \citep{sui2015safe}.  

In the standard setting, it is known that the scaling can be improved to $\sqrt{\frac{\gamma_T}{T}}$ for simple regret \citep{vakili2021optimal}, and $\sqrt{T \gamma_T}$ for cumulative regret \citep{li2022gaussian, camilleri2021high}; these improved bounds are near-optimal for common kernels such as Mat\'ern.  However, the techniques for attaining this improvement appear to be difficult to apply in the safe setting.  For instance, the approaches of \citep{li2022gaussian} and \citep{camilleri2021high} use a small number of batches (e.g., $O(\log T)$ or $O(\log \log T)$). In our setting, the safe set cannot be confidently expanded until the end of each batch, and this may be too infrequent to eventually find the safe boundary.  In view of these difficulties, we believe that attaining near-optimal $\gamma_T$ dependence in safe settings would be of significant interest for future work.

\section{Experiments} \label{sec:exps}
In this section, we present experimental results for \textsc{\sffamily M-SafeUCB}, and compare the performance to other representative algorithms for safe Bayesian optimisation \footnote{The code is available at \url{https://github.com/arpanlosalka/m-safeucb}.}. The experiments serve to (i) investigate the cumulative regret of \textsc{\sffamily M-SafeUCB} and compare against baselines, (ii) compare the boundary of the sub-level set estimated by \textsc{\sffamily M-SafeUCB} with the actual boundary, and (iii) verify that unsafe points are not sampled during the optimisation. We emphasise that the main goal of this paper is not to have our algorithm outperform or ``replace'' any baselines. Rather, our main goal is to investigate the benefits of monotonicity, particularly from a theoretical standpoint. We provide the main information regarding the functions, algorithms, and implementation here, and provide more details in Appendix B.

\paragraph{Simulated Clinical Trial:} We first evaluate the performance of \textsc{\sffamily M-SafeUCB} on a synthetic function that simulates dose-toxicity response of certain drugs. We model toxicity using the logistic function following \citep{wang2022best}, and specifically consider the following:
\begin{equation}
    f_{tox}(d, a) = \frac{1}{1+e^{-\theta da}}
\end{equation}
where $d$ and $a$ represent the drug dose (safety variable $s$) and the patient's age ($\bx$) respectively. As discussed in Section \ref{sec:intro}, we consider the scenario where both the drug toxicity and efficacy increase monotonically with increasing dosage, and thus, the task reduces to finding the \textit{maximum tolerated dose (MTD)}. We note that while the logistic function is often used to model dose-toxicity response, we incorporate the patient's age in the function here, such that the toxicity of a drug dose increases as the patient's age increases. This simulates the scenario that the \textit{MTD} for a patient with a higher age is lower than that for a patient with a lower age. We set $\theta=5$, the toxicity threshold $h=0.9$, and the range of inputs is set as $d \in [0,1], a \in [0,2]$ (after suitable scale/shift).

We use the Mat\'ern-$\frac{5}{2}$ kernel with trainable length-scale and variance parameters with log-normal priors. Based on minimal manual tuning and seeking simplicity, $\beta_t$ is set to $5$ and is kept constant throughout the optimisation (as in common in the Bayesian optimisation literature). 

\begin{figure}[t!]
  \centering
  \setlength\tabcolsep{2pt}
     \includegraphics[width=0.75\linewidth]{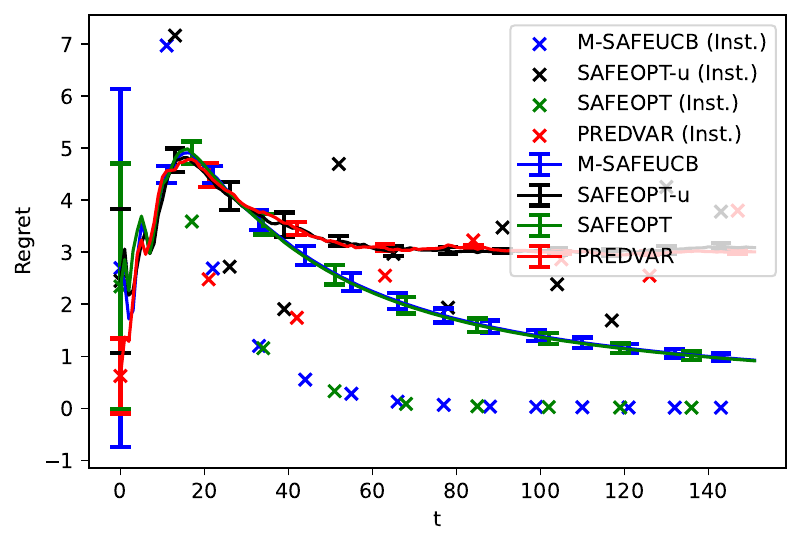} \\
     \includegraphics[width=0.75\linewidth]{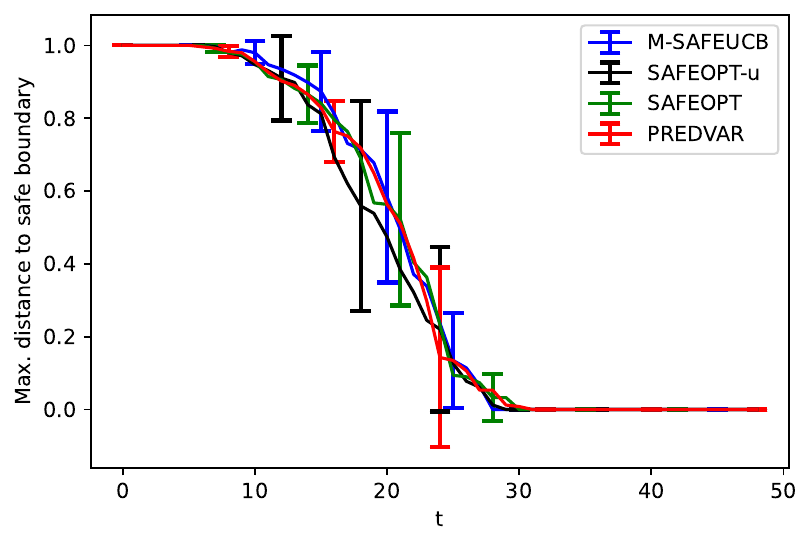}
  \caption{Results of running \textsc{\sffamily M-SafeUCB}, \textsc{\sffamily SafeOpt} and \textsc{\sffamily PredVar} on the inverted pendulum swing-up problem. The top image shows the plots for the instantaneous and average cumulative regrets incurred by the algorithms. The bottom image shows the maximum distance to the safe boundary across $x$ for the algorithms as a function of the time step.}
  \label{fig:pendulum}
\end{figure}

\begin{figure*}[t!]
  \centering
  \setlength\tabcolsep{2pt}
  \begin{tabular}{cc}
     \includegraphics[width=0.4\linewidth]{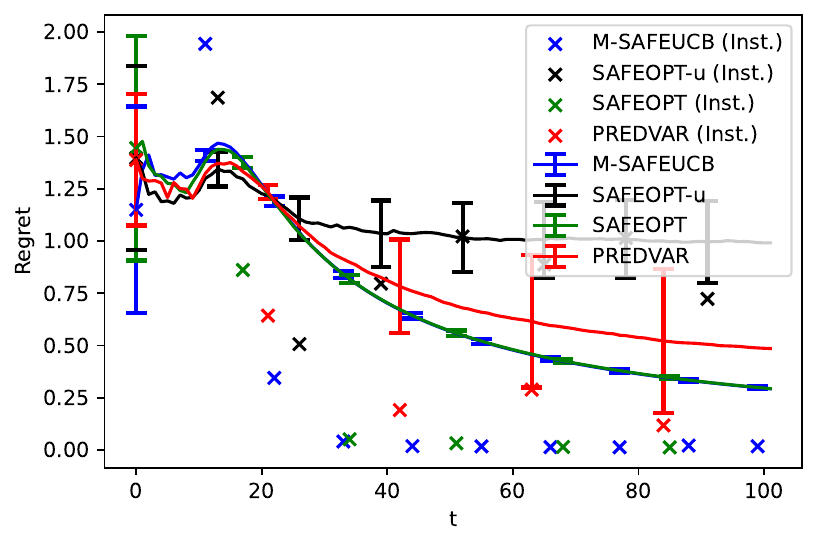} &
     \includegraphics[width=0.4\linewidth]{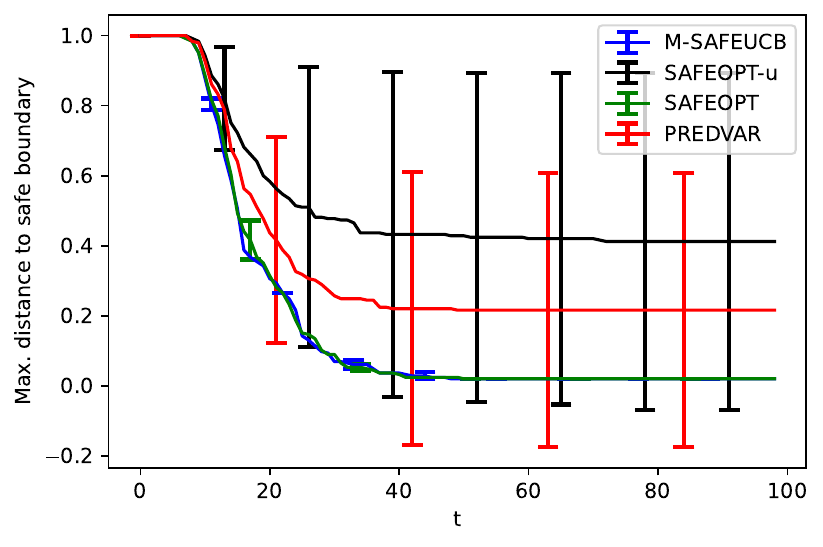}
    \end{tabular}
  \caption{Results of running \textsc{\sffamily M-SafeUCB}, \textsc{\sffamily SafeOpt} and \textsc{\sffamily PredVar} on the function $f_{syn_3}(s, \bx)$ given by \eqref{eqn:f_3D}. The left image shows the plots for the instantaneous and average cumulative regrets incurred by the three algorithms. The right image shows the maximum distance to the safe boundary across $x$ for the three algorithms as a function of the time step.}
  \label{fig:3D_input}
\end{figure*}

Fig.~\ref{fig:mgpucb_bo_safe_regret} shows the results obtained by running \textsc{\sffamily M-SafeUCB}, including the boundary of the safe set estimated, the regret incurred, and the worst-case (over $\bx$) distance to the true safe boundary as a function of time. In each case, \textsc{\sffamily M-SafeUCB} succeeds in estimating the boundary very closely without sampling unsafe points, and the instantaneous regret goes to zero, indicating sub-linear cumulative regret.

We compare the performance of \textsc{\sffamily M-SafeUCB} with the \textsc{\sffamily SafeOpt} algorithm \citep{sui2015safe}, and an active learning algorithm \textsc{\sffamily PredVar}, that simply selects the point with the highest posterior variance (among those known to be safe) in each round \citep{schreiter2015safe}. We found that all algorithms maintain the safety requirement, and are roughly equally effective at identifying the safe region. In terms of regret, however, \textsc{\sffamily PredVar} performs worse, and \textsc{\sffamily M-SafeUCB} tends to be best. 

We note here that we ran the \textsc{\sffamily SafeOpt} by using the correct estimate of the Lipschitz constant $L$, as well as an underestimate, to demonstrate the effect on performance when a proper estimate of $L$ is unavailable (\textsc{\sffamily SafeOpt}-u in Figure \ref{fig:mgpucb_bo_safe_regret}). We found that this can result in significantly degraded performance of \textsc{\sffamily SafeOpt} when $L$ is underestimated (see Appendix B for more discussion on the effect of $L$). Further, \textsc{\sffamily SafeOpt} incurs a substantially higher computation time due to the requirement of finding the set of \textit{potential expanders}.

\paragraph{Oscillating Synthetic Functions:} Next, we evaluate the performance of \textsc{\sffamily M-SafeUCB} on two functions with a more complex form of the safe boundary. The functions are defined as follows:
\begin{equation}
\begin{aligned}
    f_{syn_1}(s, x) &= (1+s)\left(1+\cos(10x)\right), \\ 
    f_{syn_2}(s, x) &= s \left( \exp(x) \sin(10x) + \sin(5x) +5 \right) / 3.
\end{aligned}
\end{equation}
Both functions are monotonically increasing in the input variable $s$, and satisfy the remaining assumptions, i.e., $\forall x \in [0,2], f(0,x) \leq h$ and $\max_{(s,x) \in \mathcal{D}}f(s,x) > h$, where the input range is set to $s \in [0,1], x \in [0,2]$ and the $h$ is set to $2$. We use the Mat\'ern-$\frac{5}{2}$ kernel as earlier, and set $\beta_t$ to $5$ for $f_{syn_1}$, and 10 for $f_{syn_2}$. 

Fig.~\ref{fig:mgpucb_bo_safe_regret} shows the results obtained by running \textsc{\sffamily M-SafeUCB}, as well as \textsc{\sffamily SafeOpt} and \textsc{\sffamily PredVar}. Similar to the earlier results, \textsc{\sffamily M-SafeUCB} achieves sub-linear regret and finds the entire safe boundary while satisfying safety constraints. \textsc{\sffamily SafeOpt} also shows similar performance when a good estimate of $L$ is available. \textsc{\sffamily PredVar} and \textsc{\sffamily SafeOpt}-u perform worse in terms of regret, as can be expected.

\paragraph{Inverted Pendulum:} We consider the \textit{inverted pendulum swing-up} problem, a classic control task from the OpenAI Gym \citep{brockman2016openai} as a representative example of problems from robotics that fit into our problem setup. The goal of this task is to apply a torque to the free end of the pendulum to swing it to an upright position, starting from a random initial position. We modify the environment to suit the assumptions of our problem statement (see Appendix B for details). The algorithm is supposed to choose the initial torque that is applied to the pendulum, and the motion is simulated for $100$ time steps. We modify the reward function as follows: (i) it equals the original reward if the pendulum does not cross the upright position, (ii) it equals zero if the pendulum reaches the upright position with zero angular velocity, and (iii) if the upright position is crossed, then we let the reward equal the angular velocity at the time of crossing.  These changes are made primarily to ensure that the resulting reward function is smooth, and the action of not applying a torque ($s=0$) is safe for all starting positions.

The goal is to maximise the reward function, while ensuring that the upright position is not crossed (thus, $h=0$), i.e., case (ii) above is the optimal one, and case (iii) is unsafe.

The experimental results of running \textsc{\sffamily M-SafeUCB} on this setup are presented in Figure \ref{fig:pendulum}, while comparing with \textsc{\sffamily SafeOpt} and \textsc{\sffamily PredVar}.  We again observe strong similarities in terms of satisfying the safety threshold and finding the safe region. In this case we find that \textsc{\sffamily SafeOpt} also closely matches \textsc{\sffamily M-SafeUCB} in terms of regret, albeit at the cost of much higher computation time.  

\paragraph{3D Input:} Similar to the earlier experiments involving synthetic functions with 2D inputs, we evaluate the performance of \textsc{\sffamily M-SafeUCB} on the following function with 3D inputs:
\begin{equation}
\label{eqn:f_3D}
    f_{syn_3}(s, \bx) = s^2 + x_1^2 + x_2^2,
\end{equation}
where $s$ denotes the safety variable and $\bx = (x_1, x_2)$ denotes the other input variable. The domain of each variable is set to $[0,1]$, and the safety threshold is set to $h = 2$, thus satisfying the assumptions that for all $\bx \in \mathcal{D_X}, f(0,\bx) \leq h$, and $\max_{(s,\bx) \in \mathcal{D}} f(s,\bx) > h$. The average cumulative regret and the instantaneous regret for \textsc{\sffamily M-SafeUCB} along with other baseline algorithms are shown in Figure \ref{fig:3D_input}.  We mostly observe similar behavior to the 2D case, except that \textsc{\sffamily PredVar} now incurs larger error bars.

\paragraph{Summary:} Overall, our experimental results illustrate that (i) explicit expansion is not necessary under our assumed monotonicity conditions, and (ii) monotonicity not only benefits our proposed algorithm, but can also benefit other baselines and safe GP exploration methods in general.

\section{Conclusion}

We have demonstrated that monotonicity with respect to a single \emph{safety variable} can have significant benefits for safe GP exploration and optimisation, including improved theoretical guarantees, algorithmic simplicity, and every safe point being reachable under mild conditions. Potential directions for future work include (i) seeking $\sqrt{\gamma_T}$ (rather than $\gamma_T$) dependence in the theoretical bounds, (ii) further studying more general scenarios with separate functions for safety and reward, (iii) determining other helpful function properties beyond monotonicity, and (iv) studying extensions to reinforcement learning settings, possibly either offline or online.

\begin{acknowledgements} 
                             This work was supported by the Singapore National Research Foundation (NRF) under grant number A-0008064-00-00.
\end{acknowledgements}

\newpage
\bibliography{losalka_421}

\appendix

\onecolumn 

{\centering
    {\huge \bf Appendix \\ [2mm]}  
}

\section{Proofs}  \label{sec:proofs}

In this section, we present the proofs for Theorem 1 and Theorem 2. 

\subsection{Proof of Theorem 1 (Regret Bound)}

By Lemma 1, with probability at least $1-\delta$, the following holds for all $(s, \mathbf{x}) \in \mathcal{D}$ and $t \geq 1$:
\begin{equation}
    |  \mu_{t-1} (s, \mathbf{x}) - f(s, \mathbf{x})| \leq 
     \beta_t \sigma_{t-1}(s, \mathbf{x})
\end{equation}
where $\mu_{t-1}(s_t, \mathbf{x})$ and $\sigma_{t-1}^2 (s_t, \mathbf{x})$ are the mean and variance of the posterior distribution.  As a special case of this fact, at each round $t \geq 1$, we have 
\begin{equation}
\label{eqn:diff_less_sigma}
\mu_{t-1}(s_t, \mathbf{x}_t)-f(s_t, \mathbf{x}_t) \leq \beta_t \sigma_{t-1}(s_t, \mathbf{x}_t).    
\end{equation}
Moreover, given the description of Algorithm 1, we have the following for all $t$:
\begin{equation}
\label{eqn:ucb_more_h}
\mu_{t-1}(s_t, \mathbf{x}_{t}) + \beta_t \sigma_{t-1}(s_t, \mathbf{x}_{t}) \geq h.
\end{equation}
(Recall that the ``safe everywhere'' step setting $s=1$ for all $\bx$ will never occur when the confidence bounds are valid, since we assume that at least one point is unsafe.)

Combining the above, we can conclude that for all $t \geq 1$, with probability at least $1-\delta$,
\begin{align}
\label{eqn:regret}
r_{t} &=h-f(s_t, \mathbf{x}_{t}) \nonumber \\
& \leq \mu_{t-1}(s_t, \mathbf{x}_t) + \beta_t\sigma_{t-1}(s_t, \mathbf{x}_{t}) - f(s_t, \mathbf{x}_{t}) \qquad \text{(by \eqref{eqn:ucb_more_h})}\nonumber  \\
& \leq 2 \beta_t \sigma_{t-1}\left(s_t, \mathbf{x}_t\right). \qquad \text{(by \eqref{eqn:diff_less_sigma})}
\end{align}
Hence, we have
\begin{equation}
R_T = \sum_{t=1}^{T} r_{t} \leq 2 \beta_t \sum_{t=1}^{T} \sigma_{t-1}\left(s_t, \mathbf{x}_{t}\right).    
\end{equation}
Now, from Lemma 4 of \citep{chowdhury2017kernelized}, $\sum_{t=1}^{T} \sigma_{t-1}\left(s_t, \bx_{t}\right)=O\big(\sqrt{T\gamma_{T}}\big)$. Furthermore, $\beta_T \leq B+R \sqrt{2\left(\gamma_{T}+1+\ln (1 / \delta)\right)}$ (since $\gamma_t$ is monotonically increasing). Hence, with probability at least $1-\delta$,
\begin{equation}
R_{T}=O\left(B \sqrt{T \gamma_{T}}+\sqrt{T \gamma_{T}\left(\gamma_{T}+\ln (1 / \delta)\right)}\right).
\end{equation}

\subsection{Proof of Theorem 2 (Identification of Safe Boundary)}

Again using Lemma 4 in \citep{chowdhury2017kernelized}, if $(s_1, \mathbf{x}_1), (s_2, \mathbf{x}_2), \dotsc, (s_T, \mathbf{x}_T)$ are the points selected by Algorithm 1, then the sum of predictive standard deviations at these points can be bounded in terms of the maximum information gain as follows:
\begin{equation}
\sum_{t=1}^T \sigma_{t-1}(s_t, \mathbf{x}_t) \leq \sqrt{4(T+2)\gamma_T}.
\end{equation}
Using the monotonicity of $\beta_t$, we deduce that for $T \geq 2$, we have
\begin{align}
\label{eqn:ci_avg}
\frac{1}{T}\sum_{t=1}^T \beta_t \sigma_{t-1}(s_t, \mathbf{x}_t) &\leq \beta_T \sqrt{4\gamma_T/T + 8\gamma_T/T^2} \nonumber \\ &\leq \beta_T\sqrt{8\gamma_T/T}.
\end{align}

\noindent Now, as per Algorithm 1, for all $\mathbf{x} \in \mathcal{D_X} \text{ and } t \leq T$, $s_t^{(\mathbf{x})}$ is one of the following:
\begin{itemize} \itemsep0ex
    \item $s_t^{(\mathbf{x})} = 0$ if it holds that $\forall s:(s, \mathbf{x}) \in \mathcal{D}, \mathrm{UCB}_{t-1}(s, \mathbf{x}) > h$;
    \item $s_t^{(\mathbf{x})}$ is undefined if it holds that $\forall s:(s, \mathbf{x}) \in \mathcal{D}, \mathrm{UCB}_{t-1}(s, \mathbf{x}) < h$;
    \item in all other cases, $s_t^{(\mathbf{x})} = \max\{s : (s, \mathbf{x}) \in \mathcal{D}, \mathrm{UCB}_{t-1}(s, \mathbf{x}) = h \}$.
\end{itemize}
Thus, we can conclude that whenever $s_t^{(\mathbf{x})}$ is defined, it satisfies
\begin{equation}
\label{eqn:monotonic_s}
    \overline{s}_{T}^{(\mathbf{x})} \geq s_t^{(\mathbf{x})} \quad \forall t \leq T.
\end{equation}
This is because $s_t^{(\mathbf{x})}$ is defined based on $\mathrm{UCB}_{t-1}$, whereas $\overline{s}^{(\mathbf{x})}_{T} = \max\{s : (s, \mathbf{x}) \in \mathcal{D}, \min_{1 \leq t \leq T} \mathrm{UCB}_{t-1}(s, \mathbf{x}) \leq h\}\}$ considers the minimum of all $\mathrm{UCB}$'s across $t$ to find the maximum $s$. 

\noindent Since \textsc{\sffamily M-SafeUCB} selects the point with the largest $\sigma_{t-1}(s, \mathbf{x})$ from the candidate set $S_t$ for $t \leq T$, we have the following whenever $s_t^{(\mathbf{x})}$ is defined:
\begin{equation}
\label{eqn:beta}
    \beta_t\sigma_{t-1}(s_t^{(\mathbf{x})}, \mathbf{x}) \leq \beta_t\sigma_{t-1}(s_t, \mathbf{x}_t) \; \forall \mathbf{x} \in \mathcal{D_X}.
\end{equation}
Next, note that $s_t^{(\mathbf{x})}$ is undefined for some $\mathbf{x} \in \mathcal{D_X}$ only if at round $t$, $\mathrm{UCB}_{t-1}(s, \mathbf{x}) < h \; \forall s: (s, \mathbf{x}) \in \mathcal{D}$. In this case, $\overline{s}_{T}^{(\mathbf{x})} =1$ by the definition. Therefore, for all $\mathbf{x} \in \mathcal{D_X}$ where this occurs for some $t$, we have $(s,\mathbf{x}) \in \hat{L}_T$ for all $s \in [0,1]$. Hence, as long as the confidence bounds are valid, we have $l_h(s, \mathbf{x}) = 0$ for such $(s,\mathbf{x})$.

\noindent For all $\mathbf{x} \in \mathcal{D_X}$ not satisfying the conditions of the previous paragraph, we have for all $t \le T$ that there exists $s \in [0,1]$ such that $\mathrm{UCB}_t(s, \mathbf{x}) \geq h$, and accordingly, $s_t^{(\mathbf{x})}$ is well-defined. In this case, we bound the maximum deviation of $f(\overline{s}_{T}^{(\mathbf{x})},\mathbf{x})$  from $h$ as follows for any $t \le T$:
\begin{align}
\label{eqn:Delta}
\Delta(\overline{s}_{T}^{(\mathbf{x})},\mathbf{x}) &:= h - f(\overline{s}_{T}^{(\mathbf{x})},\mathbf{x}) \\
& \leq h - f(s_{t}^{(\mathbf{x})},\mathbf{x}) \qquad~~~~ \label{eq:1} \text{ (by \eqref{eqn:monotonic_s} and monotonicity of } f) \\
& \leq 2\beta_{t}\sigma_{t-1}(s_t^{(\mathbf{x})},\mathbf{x}) \qquad \label{eq:2} \text{ (similar to \eqref{eqn:regret})} \\ 
& \leq 2\beta_t\sigma_{t-1}(s_t, \mathbf{x}_t), \qquad \label{eq:3}  \text{ (by \eqref{eqn:beta})}
\end{align}
provided that the confidence bounds are valid.  
\noindent 
Since this holds for all $t \le T$, we can average both sides over $t \in \{1,\dotsc,T\}$ to obtain
\begin{equation}
\begin{split}
\label{eqn:min_less_avg}
    \Delta(\overline{s}_{T}^{(\mathbf{x})}, \mathbf{x}) & \leq \frac{2}{T}\sum_{t=1}^T \beta_t \sigma_{t-1}(s_t, \mathbf{x}_t)  \\
    & \leq 2\beta_T \sqrt{8\gamma_T/T},
\end{split}
\end{equation} 
where we made use of \eqref{eqn:ci_avg}. 
Since $\hat{L}_{T} = \{(s, \mathbf{x}) \in \mathcal{D}: s \leq \overline{s}^{(\mathbf{x})}_{T}\}$, for any $(s,\mathbf{x}) \in \hat{L}_T$, we obtain $l_h(s,\mathbf{x}) = 0$ due to the validity of the confidence bounds. On the other hand, if $(s,\mathbf{x}) \notin \hat{L}_T$, there are two sub-cases to consider:
\begin{itemize} \itemsep0ex
    \item If $(s,\mathbf{x}) \notin \hat{L}_T$ and $\mathrm{LCB}_{T}(s,\mathbf{x}) > h$, then 
    \begin{equation}
    l_h(s,\mathbf{x}) = \max\{0, h - f(s,\mathbf{x})\} = 0.
    \end{equation}
    \item If $(s,\mathbf{x}) \notin \hat{L}_T$ and $\mathrm{LCB}_{T}(s,\mathbf{x}) < h < \mathrm{UCB}_{T}(s,\mathbf{x})$, then 
    \begin{align}
    l_h(s, \mathbf{x}) & = \max\{0, h - f(s,\mathbf{x})\} \\ 
    & \leq \Delta(\overline{s}_{T}^{(\mathbf{x})},\mathbf{x}) \qquad ~~~~~~\,\text{ (by }s \leq \overline{s}_{T}^{(\mathbf{x})} \text{ and monotonicity of $f$}) \\
    & \leq 2\beta_T \sqrt{8\gamma_T/T}.  \qquad \text{ (by \eqref{eqn:min_less_avg})}
    \end{align} 
\end{itemize}
\noindent Therefore, setting $\epsilon = 2\beta_T \sqrt{8\gamma_T/T}$, we have the following guarantee for \textsc{\sffamily M-SafeUCB}'s performance on the sub-level set estimation task:
\begin{equation}
\mathbb{P}\left\{\max_{s,\mathbf{x} \in \mathcal{D}} l_h(s,\mathbf{x}) \leq \epsilon \right\} \geq 1-\delta.
\end{equation} 
Substituting $\beta_T$ into the above choice of $\epsilon$ completes the proof.

\section{Details of Experiments} \label{sec:exp_details}

\paragraph{Gaussian Process Model:} For both the synthetic data and the inverted pendulum experiments, we use a Gaussian Process with Mat\'ern$\frac{5}{2}$ kernel to model the unknown function. We use the \textit{Trieste} toolbox for implementation \citep{trieste2023}, and set the length scales and variance of the kernel to be trainable. The initial variance is set by randomly sampling two points in the domain known to be safe, and computing the variance with respect to the observed function values. A log-normal prior is used for both the variance and the length scales, with a standard deviation $1$. The means for the length scales are set to $0.2$, and that for the variance is $3$. The function values returned are noiseless, while the Gaussian Process regression model assumes a low noise level of $10^{-5}$ for numerical stability. We use the Trieste library for our implementations \citep{trieste2023}.

\paragraph{Synthetic functions:} The domains of the functions $f_{syn_1}$ and $f_{syn_2}$ are set to $s \in [0,1]$ and $x \in [0,2]$. For running the algorithms, the domain is discretised into a grid with $200$ linearly spaced points in each dimension. The optimisation is run for $100$ iterations for each algorithm. Each experiment is repeated 5 times, and the mean values along with the standard deviations (via error bars) are shown.

For the function $f_{syn_3}$, the algorithms are run for $100$ iterations, with the domain discretised into a grid with $75$ linearly spaced points in each dimension. The experiments are repeated $5$ times, and the mean values and standard deviations (via error bars) of the average cumulative regret are shown.

\paragraph{Inverted Pendulum:} For this experiment, we allow the initial angle of the pendulum (denoted by $x$) to lie in $[-2\pi + \pi/36, -\pi/36]$ (where angle $0$ denotes the upright position), while the applied torque $s \in [0,1]$. The angle $\theta$ becomes positive after the pendulum crosses the upright position. We modify the reward function $f(s,x)$ as follows:
\begin{gather}
    f_n(s,x) = 
    \begin{cases}
    -\theta_n^2(s,x) -\frac{\Dot{\theta}_n^2(s,x)}{10} - \frac{s^2}{1000}, & \text{ if } \theta_n \leq 0 \\
    \Dot{\theta}_{up}(s,x)  &\text{ if } \theta_n(s,x) > 0,
    \end{cases}
    \\[2mm]
    f(s,x) = \max_{n \leq 100} f_n(s,x),
\end{gather}
where $\theta_n(s,x)$ and $\Dot{\theta}_n(s,x)$ denote the angle and angular velocity of the pendulum at the $n^{th}$ time step, and $\Dot{\theta}_{up}(s,x)$ denotes the angular velocity of the pendulum when it crosses the upright position, starting with an initial angle and torque of $x$ and $s$ respectively. Note that the time step $n$ (for simulating the motion of the pendulum) is different from the time step $t$ (denoting the optimisation iteration).

The safety threshold is set to $f(s,x) = 0$, which can only happen when both $\theta_n(s,x)$ and $\Dot{\theta}_n(s,x)$ are $0$ (since $s$ is always $0$ beyond the initial time step) for some $n \leq 100$. Thus, the safety threshold denotes the condition that the pendulum is in the upright position with a zero angular velocity, resulting in the sustenance of the upright position until the end of the episode, i.e., $n=100$.

The initial angular velocity is always set to $0$, so that our assumption that $s=0$ is a safe action is satisfied. This is because the pendulum can never swing to the upright position starting from the range of initial positions specified, unless a torque is applied. Furthermore, the initial torque is assumed to be magnified by a factor of $20$ when computing the resulting motion, resulting in the possibility of unsafe actions (torque applied, $s$) corresponding to a large fraction of starting positions (initial angle, $x$).

Similar to the experiments with synthetic data, the input domain is discretised into $200$ linearly spaced points along each dimension, and the results of running the three algorithms 5 times are presented in Figure 2.

\paragraph{Algorithm Details and Discussion:} For \textsc{\sffamily SafeOpt}, we use the version with the Lipschitz constant $L$ as proposed in the original paper \citep{sui2015safe}. We approximate $L$ by calculating the gradients for a finely discretised grid of points in the input domain in each case, and take the maximum among their magnitudes. Note that for using \textsc{\sffamily SafeOpt} in practice, $L$ needs to be tuned alongside $\beta_t$ as a hyperparameter. We consider the ``best case'' here for \textsc{\sffamily SafeOpt}, where a close approximation of the original Lipschitz constant for the unknown function is known to the algorithm. For the version of the algorithm using an underestimate of $L$ in the experiments, we reduce the estimated $L$ by a factor of $2$ to $5$. Further, we use the techniques discussed in Section 4 of \citep{berkenkamp2017safe} to reduce the computation cost of \textsc{\sffamily SafeOpt}. Despite these optimisations, we found that \textsc{\sffamily SafeOpt} can incur more than ten times the computation cost of \textsc{\sffamily M-SafeUCB} in our experiments, and this difference increases with increasing input dimension.

As discussed in Section 4 of \citep{sui2015safe}, we solely use the confidence intervals for guaranteeing safety, and only use the Lipschitz constant for finding potential expanders. It is important to note here that using the Lipschitz constant for determining the safe set $S_t$ further increases the dependence of \textsc{\sffamily SafeOpt} on the value of $L$, and can lead to degradation in performance due to over-cautiousness when $L$ is overestimated. Thus, we avoid this version of the algorithm in our experiments. We also investigated the modified \textsc{\sffamily SafeOpt} algorithm suggested in \citep{berkenkamp2017safe} that avoids the dependence on $L$ altogether, but we found it to be substantially more time consuming to run.

We also note here that in the version of \textsc{\sffamily SafeOpt} used in our experiments as described above, overestimating $L$ essentially makes the algorithm behave very similar to \textsc{\sffamily M-SafeUCB}, since the number of points included in the set of potential expanders is small (or even zero) due to over-cautiousness, while the set of maximisers remains unaffected since $S_t$ is determined only using the confidence intervals of the GP. This leads to wasteful computation compared to \textsc{\sffamily M-SafeUCB} leading to a greatly increased running time for obtaining a similar performance, thus also showcasing the benefits of using \textsc{\sffamily M-SafeUCB} over \textsc{\sffamily SafeOpt} for the problem setup under consideration.

For the \textsc{\sffamily PredVar} algorithm, we consider the variance of all points in the domain with $s=0$ (since these are known to be safe), as well as the points that can be guaranteed to be safe based on $\mathrm{UCB}_{t-1}$ at time step $t$, and choose the one with the highest variance.

We also note that \textsc{\sffamily M-SafeUCB} is similar in spirit to the (\textsc{\sffamily SafeUCB}) baseline \citep{sui2015safe}, which simply maximises the UCB among all points that are known (with high probability) to be safe.  However, doing this naively would lead to focusing on a small region of the $\bx$ space and ignoring the rest. \textsc{\sffamily M-SafeUCB} overcomes this by using the maximum-variance rule.

\section{Further Discussion}

\subsection{Discussion on $\mathcal{D_X}$ Dependence} \label{sec:domain_size}

\begin{figure}
    \begin{centering}
        \includegraphics[width=0.5\columnwidth]{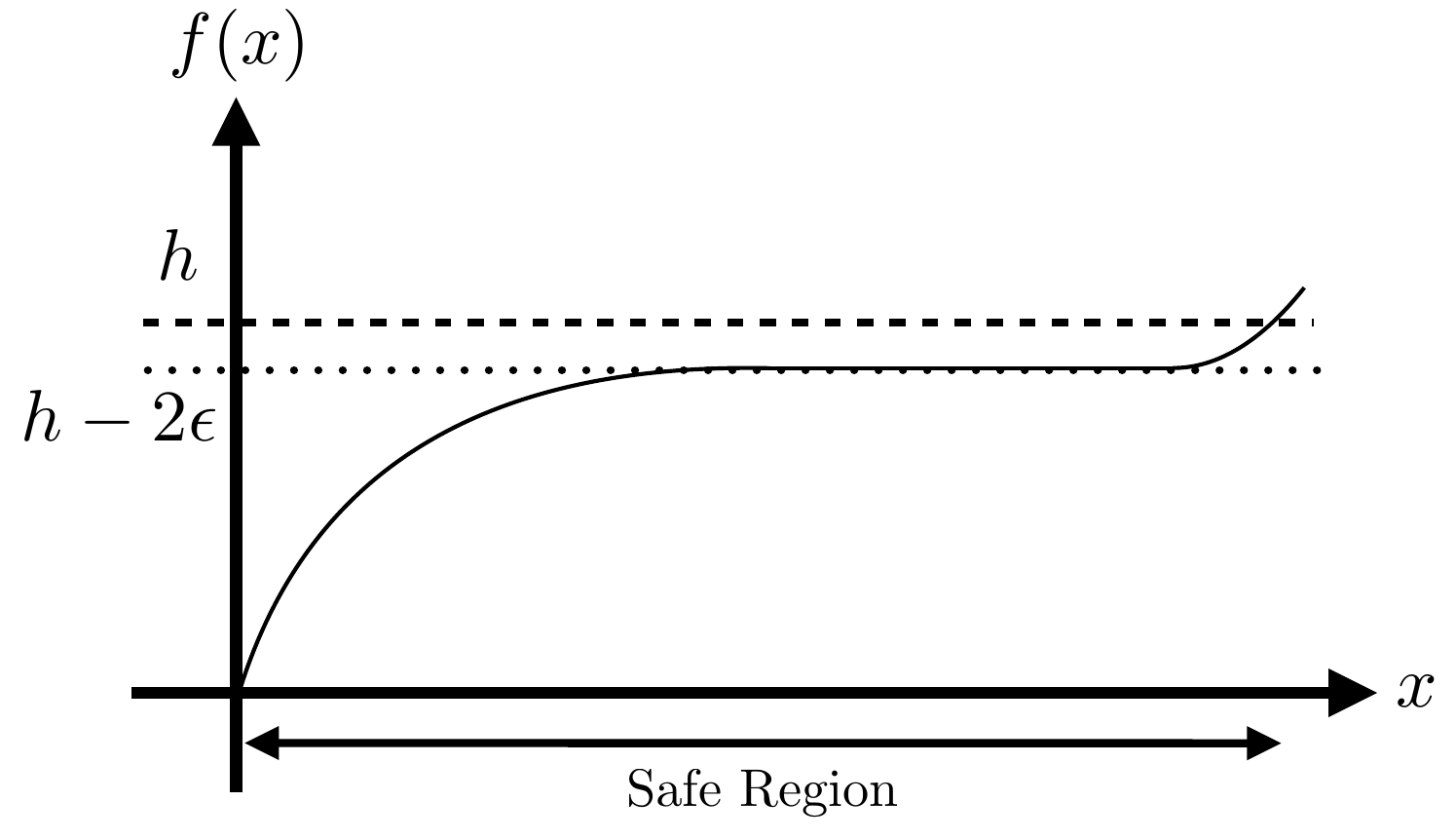}
        \par 
    \end{centering}
    
    \caption{Example of a 1D function where expanding the known safe set (i.e., the points with $f(x) \le h$) may be slow. \label{fig:Safe1D}}
\end{figure}

To get some intuition on why a linear dependence on the domain size may arise for algorithms such as \textsc{\sffamily SafeOpt} (as discussed in Section 4), consider the function shown in Figure \ref{fig:Safe1D}.  Once the function reaches $h - 2\epsilon$, it may become very difficult to use the confidence bounds and Lipschitz constants (as \textsc{\sffamily SafeOpt} uses) to determine whether it is still safe to move further to the right.  One can imagine that an algorithm ends up sampling every $x$ (or at least most $x$) even if $[0,1]$ is discretised rather finely, particularly if the Lipschitz constant is over-estimated.

On the other hand, we highlight some potential weaknesses of \textsc{\sffamily SafeOpt} via two perspectives as follows:
\begin{itemize}
    \item[(i)] If the domain is quantised very finely, then one should only expect a number of samples depending on $\frac{L}{\epsilon^2}$, rather than $\frac{|\mathcal{D_X}|}{\epsilon^2}$.  This is because once a given point with $f(x)=h-2\epsilon$ has its function value known accurately (say, to within $0.5\epsilon$), one should be able to certify the entire surrounding region of width $O(1/L)$ as safe, rather than only the next point to the right.
    \item[(ii)] One can attain a guarantee with $T$ having $\frac{|\mathcal{D_X}|}{\epsilon^2}$ or even $\frac{L}{\epsilon^2}$ dependence (up to logarithmic factors) using a fairly trivial algorithm: Repeatedly sample all (known) safe points until their function values are known to within $0.5\epsilon$ using basic concentration bounds, then expand the safe set using the Lipschitz constant, then return to repeated sampling (only for points not yet sampled), and so on.  (Logarithmic terms would then arise from applying the union bound.)  The resulting guarantee would even further improve \textsc{\sffamily SafeOpt}'s guarantee due to omitting $\beta_T \gamma_T$ on the left-hand side.
\end{itemize}
Despite these limitations, we note that \textsc{\sffamily SafeOpt} has been an important and highly influential algorithm since its introduction, and the above discussion is only meant to highlight that its theoretical guarantees, while valuable, may leave significant room for improvement in certain scenarios.

\subsection{Computational Considerations}
\label{sec:computation}

As stated, Algorithm 1 involves an explicit loop over all $\mathbf{x} \in \mathcal{D_X}$.  This is feasible when the domain size is small, and we adopted it in our experiments.  However, such an approach may become infeasible for large or continuous domains.  In such cases, one may need to rely on approximations or alternative methods, some of which we briefly discuss here.

First, if the domain is continuous, then one could rely on any \emph{constrained black-box (non-convex) optimisation} solver to minimise the posterior variance subject to the UCB being at most $h$.  For commonly-used kernels, the posterior variance and UCB are differentiable, which can facilitate this procedure.  Moreover, to handle the possibility of points with $s=0$ being selected, a second constrained black-box search could be performed over all $(0,\bx)$ subject to the UCB being \emph{at least} $h$.   The final selected point would then be the higher-variance one among the two points identified.

If no suitable black-box solver is available, or if the domain is discrete but large, then a simple practical alternative is as follows.  Instead of performing a full optimisation of the acquisition function, one can randomly select a moderate number of $\bx$ points at random (e.g., 500 or 1000) and only optimise over those.  Due to the randomness, $\bx$'s throughout the entire domain will then be considered regularly with high probability.  Moreover, the efficiency could potentially be improved by ruling out certain regions early (e.g., when $s=1$ is known to be safe).  Note, however, that we do not claim any theoretical guarantees under these variations of the algorithm.

\subsection{Discussion of \cite{amani2021regret}} \label{sec:amini}

As we discussed in Section 1, the approach of \citep{amani2021regret} is based on first expanding the safe set using sufficiently many samples within an initial seed set.  To highlight a limitation of this approach for certain kernels with infinite-dimensional feature spaces, consider the Mat\'ern kernel, and suppose that the initial seed set includes a large fraction of the domain, but the function value is zero within that entire set.  Since compactly supported ``bump'' functions are in the Mat\'ern class \citep{bull2011conv}, the function may contain both positive and negative bumps outside the seed set, some of which are safe and some of which are not.  (Here we only assume that $f(\cdot) = 0$ is safe.)  Since the function is zero within the seed set, there is no way that its samples can distinguish between these two cases.

In contrast, for finite-dimensional feature spaces (e.g., the linear or polynomial kernel) even samples within a small seed set can indeed be sufficient to accurately learn the entire function.  Finally, for the infinite-dimensional case with very rapidly decaying eigenvalues (e.g., SE kernel), the situation is somewhere in between the preceding examples; in particular, compactly supported functions are not in the RKHS.  In such scenarios, the approach of \citep{amani2021regret} may be feasible, though the precise details become somewhat complicated; certain results for infinite-dimensional settings are given in \citep{amani2021regret} accordingly.



\end{document}